\RequirePackage{fix-cm}
\documentclass[twocolumn]{svjour3}         
\smartqed 
\usepackage{graphicx}
\usepackage{natbib}
\usepackage[pagebackref=false,breaklinks=true,citecolor=blue,colorlinks,linkcolor=red,bookmarks=false,urlcolor=blue]{hyperref}

\usepackage{array}
\usepackage{multirow}
\newcolumntype{H}{>{\setbox0=\hbox\bgroup}c<{\egroup}@{}}
\usepackage{amsmath}
\usepackage{bm}
\usepackage{amsfonts}

\usepackage{hyperref}
\usepackage{graphicx}
\usepackage{amssymb}

\usepackage[misc]{ifsym} 

\usepackage{bbm}
\usepackage{dsfont}
\usepackage{pifont}

\usepackage{microtype}

\usepackage{booktabs}

\usepackage{subfig}

\usepackage[dvipsnames, table]{xcolor}
\definecolor{forestgreen}{rgb}{0.13,0.55,0.13}

\usepackage{xspace}
\newcommand{\onedot}{\ifx\@let@token.\else.\null\fi\xspace}
\newcommand{\eg}{\emph{e.g}\onedot} 

\newcommand{\ie}{\emph{i.e}\onedot}

\begin{document}
\sloppy 

\title{Centerness-based Instance-aware Knowledge Distillation with Task-wise Mutual Lifting for Object Detection on Drone Imagery}

\author{Bowei Du$^{1,2,\dag}$ \and Zhixuan Liao$^{1,2,\dag}$ \and Yanan Zhang$^{1,2}$ \and  Zhi Cai$^{1,2}$ \and  Jiaxin Chen$^{2}$ \and Di Huang\textsuperscript{\rm 1,2,3,\Letter} 
}

\authorrunning{Bowei Du {\it et al.}}
\titlerunning{Centerness-based Instance-aware Knowledge Distillation with Task-wise Mutual Lifting}

\institute{
Bowei Du\\ \email{boweidu@buaa.edu.cn}
\\~\\ 
Zhixuan Liao\\ \email{zxliao2000@buaa.edu.cn}
\\~\\ 
Yanan Zhang\\ \email{zhangyanan@buaa.edu.cn}
\\~\\ 
Zhi Cai\\ \email{caizhi97@buaa.edu.cn}
\\~\\ 
Jiaxin Chen\\ \email{jiaxinchen@buaa.edu.cn} 
\\~\\ 
Di Huang\\ \email{dhuang@buaa.edu.cn} \\~\\
$^1$ State Key Laboratory of Software Development Environment, Beihang University, Beijing, China \\~\\
$^2$ School of Computer Science and Engineering, Beihang University, Beijing, China \\~\\
$^3$ Hangzhou Innovation Institute, Beihang University, Hangzhou, China
\\~\\
$^{\dag} $indicates equal contribution.
\\~\\
\textsuperscript{\rm \Letter} refers to the corresponding author.
}

\date{Received: date / Accepted: date}

\maketitle

\begin{abstract}
Developing accurate and efficient detectors for drone imagery is challenging due to the inherent complexity of aerial scenes. While some existing methods aim to achieve high accuracy by utilizing larger models, their computational cost is prohibitive for drones. Recently, Knowledge Distillation (KD) has shown promising potential for maintaining satisfactory accuracy while significantly compressing models in general object detection. Considering the advantages of KD, this paper presents the first attempt to adapt it to object detection on drone imagery and addresses two intrinsic issues: (1) low foreground-background ratio and (2) small instances and complex backgrounds, which lead to inadequate training, resulting insufficient distillation. Therefore, we propose a task-wise Lightweight Mutual Lifting (Light-ML) module with a Centerness-based Instance-aware Distillation (CID) strategy. The Light-ML module mutually harmonizes the classification and localization branches by channel shuffling and convolution, integrating teacher supervision across different tasks during back-propagation, thus facilitating training the student model. The CID strategy extracts valuable regions surrounding instances through the centerness of proposals, enhancing distillation efficacy. Experiments on the VisDrone, UAVDT, and COCO benchmarks demonstrate that the proposed approach promotes the accuracies of existing state-of-the-art KD methods with comparable computational requirements. Codes will be available upon acceptance.

\end{abstract}

\section{Introduction}
\label{sec:intro}

Unmanned Aerial Vehicle (UAV) equipment has played a pivotal role in numerous applications such as traffic management and security surveillance, due to its commendable flight capabilities and user-friendly operation, which stimulates a surge in the necessity for vision tasks, in particular object detection. Although general object detection has largely advanced during the last decade with sophisticated models and increasing accuracies delivered, relatively large computational requirements make it difficult to adapt them to resource-constrained drone hardware, thereby posing a significant challenge in developing effective and efficient detectors on drone imagery.

\begin{figure*}[!t]
    \centering
    \resizebox{0.95\linewidth}{!}{
        \subfloat[]{
            \centering
            \includegraphics[width=0.29\textwidth,keepaspectratio]{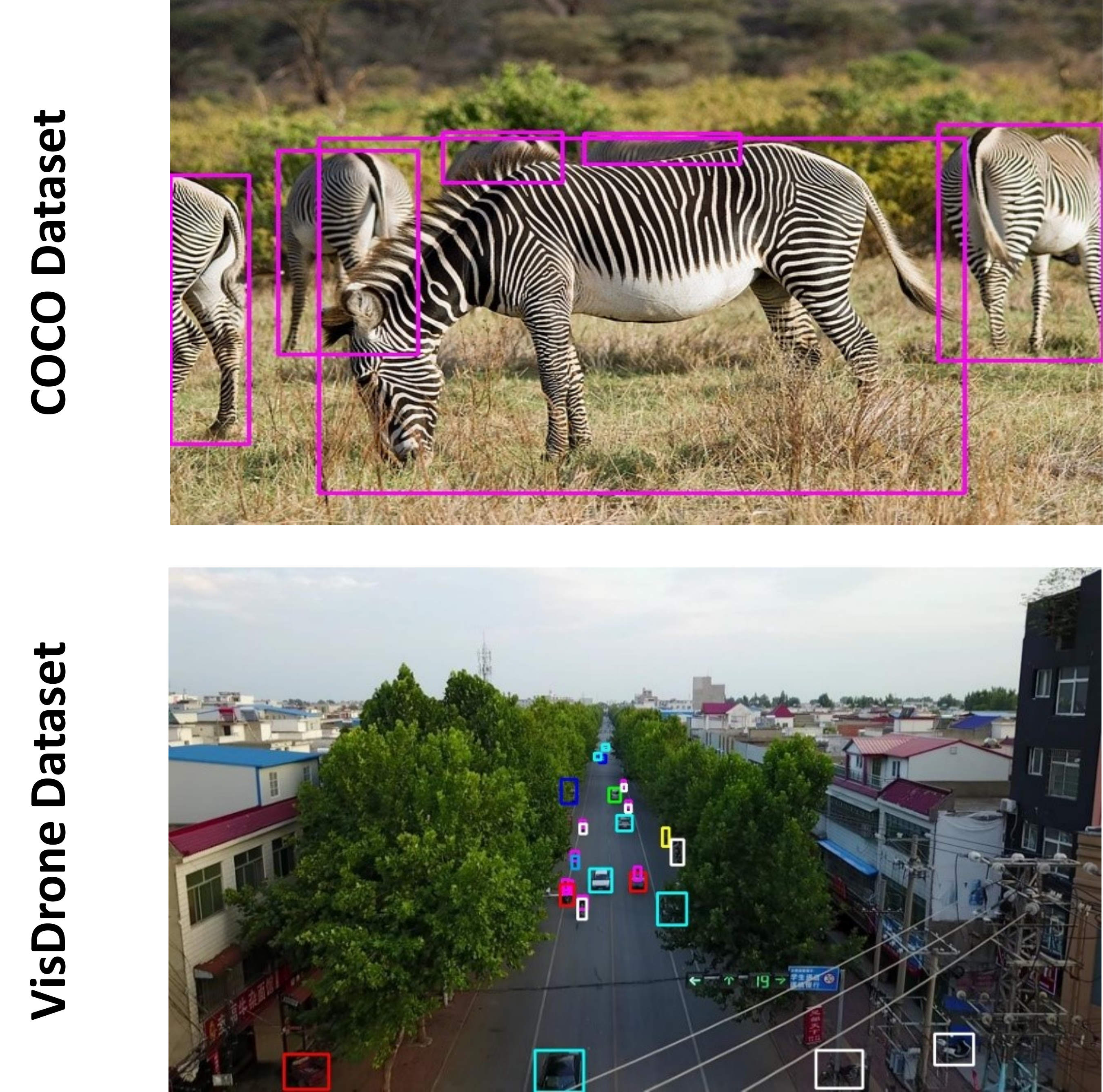}
        }
        \rule[0.5ex]{0.5pt}{\dimexpr.28\linewidth}
        \subfloat[]{
            \centering
            \includegraphics[width=0.65\textwidth,keepaspectratio]{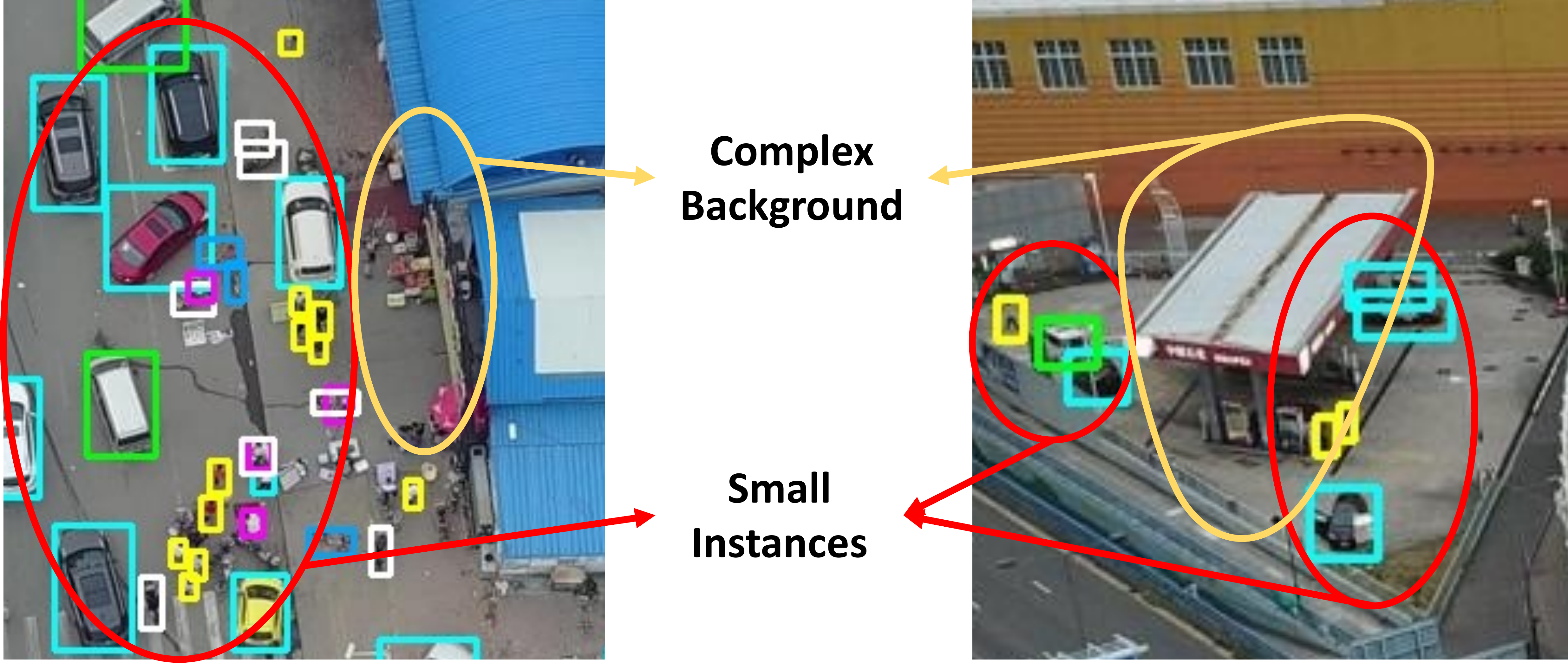}
        }
    }
    \caption{Challenges to object detection on drone imagery (VisDrone): (a) low foreground-background ratio and (b) small instances and complex backgrounds.}
    \label{fig:challenge}
\end{figure*}

A number of efforts have been made to address the trade-off between accuracy and efficiency of detectors on drone images through model compression techniques, including lightweight network methods~\citep{tphyolov52021,fasterx2022,yolodrone2023}, sparse convolution methods~\citep{QueryDet2022,ceasc2023} and network pruning methods~\citep{groupfisherprune2021,slimyolov32019}. Nevertheless, lightweight network methods design elaborate modules to improve the performance without reducing much computational load, while network pruning methods tend to disregard essential calculations on small objects which are dominant in drone images, thus leading to a sub-optimal accuracy. Additionally, sparse convolution methods show a better speed-accuracy trade-off by implementing constrained sparse convolutions, albeit at the cost of compromising the generalizability of the model on drone platforms.

Recently, Knowledge Distillation (KD) has emerged as a promising alternative to model compression, offering largely reduced training and deployment costs while maintaining competitive performance, and has proved effective in general object detection. KD typically follows a student-teacher framework, where knowledge of the larger teacher is distilled and transferred to the smaller student, thus generating a model of a compressed size with an enhanced accuracy. 

Unfortunately, it is not so straightforward to adapt KD to detectors on drone images, and two major challenges remain. First, as illustrated in Fig.~\ref{fig:challenge} (a), drone images typically exhibit a low ratio between foreground and background, which limits the supervisory information conveyed by the teacher model in the foreground area. This limited supervision leads to insufficient distillation, resulting in a more pronounced performance gap between the teacher and student models. Second, when applying KD in object detection, a common practice is to extract additional valuable regions complementing positive samples for teacher supervision reinforcement. The strategies mainly include foreground-background region~\citep{defeat2021,fgfi2019,FGD2022}, teacher-student comparison~\citep{gid2021}, and IoU-based selection~\citep{ld2022}. However, regarding the case of drone imagery, instances are often small and backgrounds are usually complicated as shown in Fig.~\ref{fig:challenge} (b), both of which make it more difficult to extract additional regions, further limiting the supervisory information, and thus incur an inadequate distillation effect.

To address the aforementioned challenges, this paper proposes a novel approach with a task-wise Lightweight Mutual Lifting (Light-ML) module and a Centerness-based Instance-aware Distillation (CID) strategy for drone images. To mitigate the insufficient supervision caused by the low foreground-to-background ratio, the Light-ML module is designed to mutually harmonize the classification and localization branches through an innovative module consisting of channel shuffling and convolution. It enhances the input information for both branches, and integrates teacher supervision clues across different tasks during the back-propagation stage, thereby amplifying the impact of KD, and effectively reducing the performance gap between the teacher and student models. For the issue of insufficient distillation due to difficulties in extracting additional information from small instances in complex background, the CID strategy generates valuable regions around instances based on the centerness of the prediction of each anchor box or anchor point, which enables the smooth and flexible estimation of informative regions surrounding objects, particularly concerning small instances. 

The contribution of our work lies in three-fold:
\begin{itemize}
	\item We propose a novel KD approach to object detection in particular for drone images. To the best of our knowledge, this is the first attempt to introduce KD techniques for model compression to build detectors for drone imagery.
	\item We design a Lightweight Mutual Lifting (Light-ML) module and a centerness-based instance-aware distillation (CID) strategy to address the inherent challenges of applying KD to drone imagery detection. These components enrich the supervision information during distillation, thereby enhancing the distillation effect.
	\item We make extensive evaluation on three public benchmarks (VisDrone, UAVDT, and COCO) with various detection pipelines (\emph{e.g.} GFL v1 and ATSS) and achieve the state-of-the-art accuracy, showing the great potential of KD in this field.
\end{itemize}

\section{Related Work}

\subsection{General Object Detection}

Most CNN-based general object detection methods are categorized into multi-stage detectors~\citep{Faster-RCNN2015,MaskRCNN2017,CascadeRCNN2018}, one-stage anchor-based detectors~\citep{RetinaNet2017,ATSS2020,GFLv12020,GFLv22021} and one-stage anchor-free detectors~\citep{centernet2019,FSAF2019,FCOS2019}. Multi-stage detectors first generate proposal regions and subsequently refine classification and regression results in the next stage. Conversely, one-stage detectors directly classify and locate objects within the entire feature. Anchor-based detectors utilize prior anchor boxes to identify potential proposals or predictions, while anchor-free detectors alleviate complicated computation associated with anchor boxes by leveraging key points of the object \eg centerness constraints. 

\subsection{Drone Images Object Detection}

Contemporary studies mainly concentrate on promoting detection accuracy in drone images and usually employ a coarse-to-fine framework, which first adopts a coarse network to approximate the locations of regions harboring densely distributed objects and then a fine network to detect small objects within these identified regions. ClusDet~\citep{ClusDet2019} integrates a sub-network to generate coarse regions and proposes a scale estimation network for better fine detection. DMNet~\citep{DMNet2020} introduces density map into coarse estimating and utilizes sliding window to get minimal regions. UFPMP-Det~\citep{ufpmpdet2022} employs a mosaic-based approach to merge estimation results into a unified image and employs a Multi-Proxy Detection Network to improve the classification accuracy. Focus\&Detect~\citep{FocusAndDetect2022} introduces the Gaussian Mixture Model for coarse regions estimation and proposes an incomplete box suppression algorithm to avoid overlapping regions.

In contrast to the inefficiency of the coarse-to-fine detection methodology, various methods based on lightweight detection models are adopted. Slim-YOLOv3~\citep{slimyolov32019} introduces network slimming~\citep{networkslimming2017} into YOLOv3~\citep{yolov32018} while TPH-YOLOv5~\citep{tphyolov52021} and FasterX~\citep{fasterx2022} introduce attention-based modules into YOLOv5~\citep{yolov52020} and YOLOX~\citep{yolox2021} to improve the detection accuracy on drone images respectively. QueryDet~\citep{QueryDet2022} and CEASC~\citep{ceasc2023} introduce sparse convolution into detectors as the low foreground ratio on drone images. Nevertheless, these methods encounter limitations concerning their acceleration capabilities and the inherent complexity of drone images, making it challenging to achieve an effective trade-off between speed and efficiency.

\subsection{Knowledge Distillation for Object Detection}

Knowledge Distillation (KD) serves as an efficient model compression method to train lightweight student models through extra supervised knowledge generated from more powerful teacher models. KD has been first introduced into the classification task~\citep{kd2015, fitnet2014, rkd2019, irg2019, dkd2022, btbs2022} and demonstrated its efficacy. On the object detection task, KD can be broadly divided into feature-based distillation~\citep{defeat2021, FGD2022, pkd2022, mgd2022} and logit-based distillation~\citep{ld2022, bckd2023}. Feature-based distillation focuses on multi-scale intermediate features from the FPN output. FGD~\citep{FGD2022} separates the foreground and background features, employing attention-based supervision to enhance information extraction within focal areas. PKD~\citep{pkd2022} leverages the Pearson Correlation Coefficient to focus on the relational information. MGD~\citep{mgd2022} introduces a masked image modeling structure, establishing a general feature-based distillation approach. GKD~\citep{gkd2024} leverages gradient information to highlight and distill the most impactful feature. Logit-based distillation focuses on logits from classification and localization tasks. LD~\citep{ld2022} leverages the probability distribution of bounding boxes proposed by GFL~\citep{GFLv12020}, applying Kullback-Leible divergence for distillation, and proposes an IoU-based Region Weighting method. BCKD~\citep{bckd2023} formulating classification logit maps as multiple binary classification maps, mitigating inconsistency issues. These methods have not considered the performance gap between teacher-student models and information extraction on small objects, resulting in restricted performance in drone images. In contrast, our method introduces lightweight student feature lifting with instance-aware information extraction to achieve better accuracy.

\begin{figure*}[t]
    \centering
    \resizebox{0.9\linewidth}{!}{
        \includegraphics[width=1.0\linewidth,keepaspectratio]{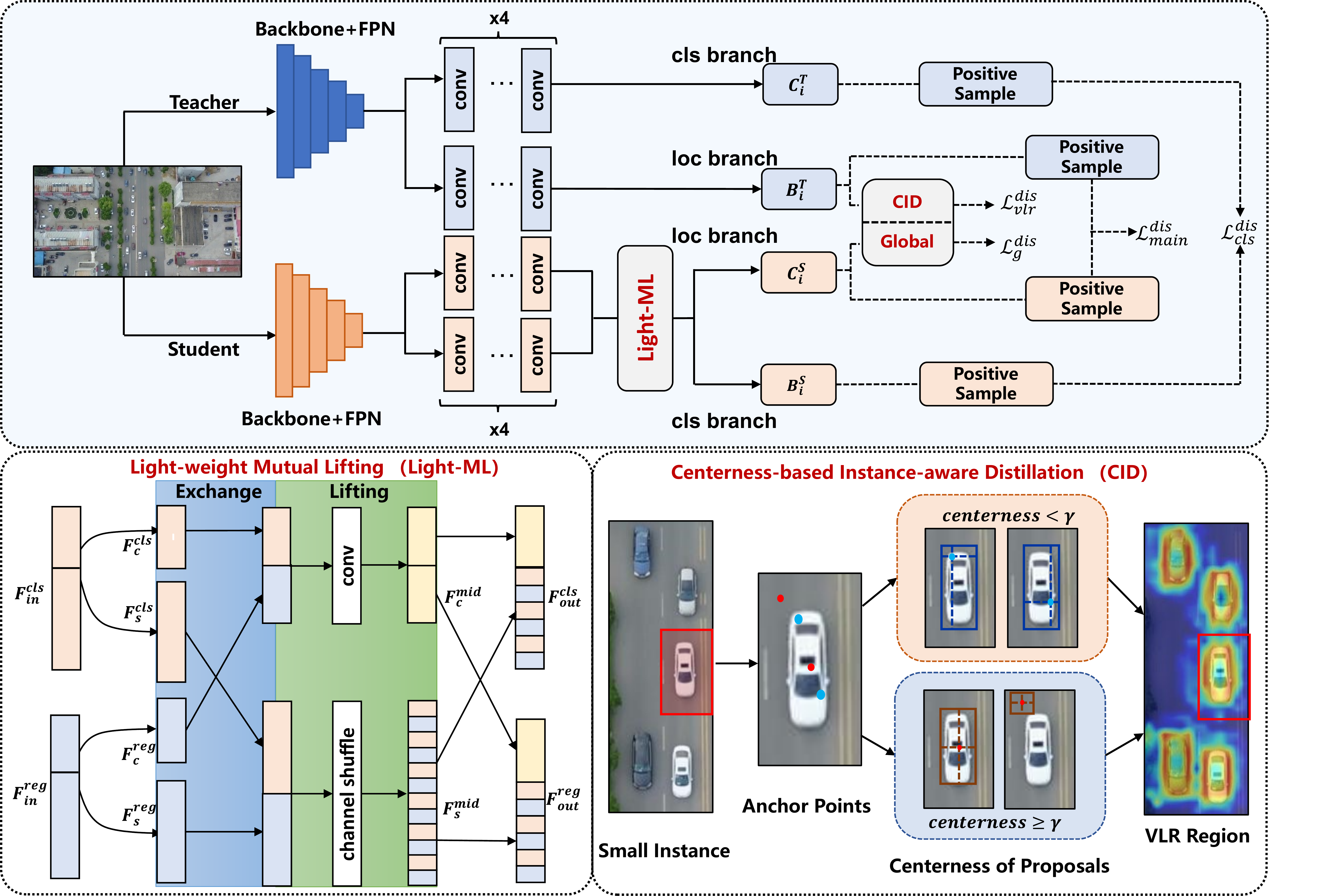}
    }
   
    \caption{Framework of our proposed distillation approach. Lightweight Mutual Lifting (Light-ML) mechanism is integrated into the detection heads of the student model for feature lifting and Centerness-based Instance-aware Distillation (CID) introduces an adaptive knowledge weighting algorithm for focal distillation combined with global distillation, enabling the extraction of additional supervision information from the teacher models. Notably, our proposed method can be applied to existing logit-based distillation approaches easily.} 
   
    \label{fig:framework}
\end{figure*}

\section{The Proposed Method}

We propose a knowledge distillation framework as shown in Fig.~\ref{fig:framework}, which employs a logit-based distillation pipeline and integrates the Lightweight Mutual Lifting module into the detection head to mitigate the inadequate training in student model by lifting the distillation supervision information. Additionally, we incorporate the Centerness-based Instance-aware Distillation method to facilitate the generation of optimal valuable regions for information extraction. 

In this section, we first briefly review two primary types of KD pipelines. Next, we present the Lightweight Mutual Lifting module and the Centerness-based Instance-aware Distillation method, respectively.

\subsection{Preliminaries}
\label{sec:method_preliminaries}

Prevalent object detection models typically comprise a backbone network and task-specific detection heads. Given an input image $\bm{X}_{input}$, the detector leverages the backbone network along with Feature Pyramid Network (FPN) to extract feature information $\bm{F}_1,...,\bm{F}_N$ across multiple scales. Subsequently, the task-specific detection heads process $\bm{F}_i$ individually for classification and localization tasks in different scales, and yield the detection results $\bm{\mathcal{C}}_{i} \in \mathbb{R}^{n \times C}$ and $\bm{\mathcal{B}}_{i} \in \mathbb{R}^{n \times 4}$ or $\bm{\mathcal{B}}_{i} \in \mathbb{R}^{n \times 4D}$, where $n$ represents the number of anchors in this scale, $C$ represents the number of target classes, and $D$ represents the number of regression distributions~\citep{GFLv12020, ld2022}. 

For distillation on object detection, feature-based approaches focus on distilling intermediate features $\bm{F}_i$:
\begin{equation}
    \mathcal{L}_{dis}=\frac{1}{N} \sum_i{\mathbb{\bm{I}}_{feat}L_{feat}(\bm{F}^T_i, \bm{F}^S_i)}
\end{equation}
where $\mathbb{I}$ represents distillation information weighting result, $L$ represents distilation loss function, $\bm{F}^T_i, \bm{F}^S_i$ are intermediate features from teacher and student detector. And logit-based approaches focus on distilling detection results $\bm{\mathcal{C}}_{i}$ and $\bm{\mathcal{B}}_{i}$:
\begin{equation}
    \mathcal{L}_{dis}=\frac{1}{N} \sum_i{\bm{\mathbb{I}}_{cls}L_{cls}(\bm{\mathcal{C}}^T_i,\bm{\mathcal{C}}^S_i)+\bm{\mathbb{I}}_{reg}L_{reg}(\bm{\mathcal{B}}^T_i,\bm{\mathcal{B}}^S_i)}
\end{equation}
where $\bm{\mathcal{C}}^T_i, \bm{\mathcal{C}}^S_i$ are classification results from teacher and student detector, $\bm{\mathcal{B}}^T_i, \bm{\mathcal{B}}^S_i$ are localization results from teacher and student detector.

\subsection{Lightweight Mutual Lifting}
\label{sec:method_light-ml}

According to~\citep{efficacykd2019,improvedkd2020,btbs2022}, we attribute the degradation in distillation on drone imagery to the capability gap between teacher and student models. Specifically, the low foreground ratio in drone imagery results in fewer positive regions, leading to insufficient supervisory information during training. Additionally, the student model's backbone exhibits limitations in information extraction while the parameter ratio within detection heads increases as the backbone size diminishes, leading to inadequate training under limited supervision, resulting in degraded distillation performance. To intuitively demonstrate the inadequate training issue in drone image detection task, we conduct experiments as described in the Section~\ref{sec:ablation_on_lightml}.

To address this issue, a common practice is to reduce the number of parameters in the detection head. However, it weakens the student model's representation and further widen the capability gap between the teacher and student models. Therefore, we propose a method that enables adequate training without compromising the student model's representation.

Inspired by ShuffleNet~\citep{ShufflenetV12018}, which uses cross-group information lifting operations such as channel shuffle, to integrate information during forward propagation and enhance the model's representation, we apply a similar concept. Additionally, during backward propagation, cross-group information flow provides supplementary supervisory information to each group through the back-propagated gradient.

Therefore, to achieve adequate training without compromising the student model's representation, we propose the Lightweight Mutual Lifting module (Light-ML), which enhances the performance of inadequately trained detection heads by lifting features from both the classification and localization branches.

As displayed in Fig.~\ref{fig:framework}, we utilize the results from feature extraction subnets of student models as inputs, namely $\bm{F}^{cls}_{in}\in \mathbb{R}^{N\times C\times H\times W}$ and $\bm{F}^{reg}_{in}\in \mathbb{R}^{N\times C\times H\times W}$. While directly applying convolutional operators can effectively achieve cross-task information lifting, the increased number of feature channels leads to higher computational load. On the other hand, using feature processing guided by attention mechanism may be constrained in both operational speed and deployment efficiency. Consequently, we introduce a CSP-wise structure~\citep{cspnet2020} to reduce module computations. Specifically, we split the input features into the convolutional part as $\bm{F}^{cls}_{c}\in \mathbb{R}^{N\times kC\times H\times W}$, $\bm{F}^{reg}_{c}\in \mathbb{R}^{N\times kC\times H\times W}$ and the shuffling part as $\bm{F}^{cls}_{s}\in \mathbb{R}^{N\times (1-k)C\times H\times W}$, $\bm{F}^{reg}_{s}\in \mathbb{R}^{N\times (1-k)C\times H\times W}$, where $k\in [0,1]$ denotes the division ratio.

Firstly, we introduce channel shuffle~\citep{ShufflenetV12018} between $\bm{F}^{cls}_{s}$ and $\bm{F}^{reg}_{s}$ to facilitate the exchange of feature information between these two branches, thus enabling feature fusion without incurring additional computational costs. And we further apply convolution on $\bm{F}^{cls}_{c}$ and $\bm{F}^{reg}_{c}$ to achieve enhanced feature lifting:
\begin{equation}
\bm{F}^{mid}_{s}=c\textrm{-}shuffle(concat(\bm{F}^{cls}_{s}, \bm{F}^{reg}_{s}))
\end{equation}
\vspace{-1.5\baselineskip}
\begin{equation}
\bm{F}^{mid}_{c}=conv(concat(\bm{F}^{cls}_{c}, \bm{F}^{reg}_{c}))
\end{equation}
where $c\textrm{-}shuffle$ indicates the channel shuffle function. $\bm{F}^{mid}_{s} \in \mathbb{R}^{N\times 2(1-k)C\times H\times W}$ and $\bm{F}^{mid}_{c} \in \mathbb{R}^{N\times 2kC\times H\times W}$ are then evenly split into $\bm{F}^{'cls}_{s}$, $\bm{F}^{'reg}_{s}$ and $\bm{F}^{'cls}_{c}$, $\bm{F}^{'reg}_{c}$. Finally, we derive $\bm{F}^{cls}_{out} \in \mathbb{R}^{N\times C\times H\times W}$ and $\bm{F}^{reg}_{out} \in \mathbb{R}^{N\times C\times H\times W}$ by concatenating the obtained features, as illustrated in Fig.~\ref{fig:framework}.
\begin{equation}
    \bm{F}^{cls}_{out}=concat(\bm{F}^{'cls}_{c}, \bm{F}^{'cls}_{s})
\end{equation}
\vspace{-1.5\baselineskip}
\begin{equation}
    \bm{F}^{reg}_{out}=concat(\bm{F}^{'reg}_{c}, \bm{F}^{'reg}_{s})
\end{equation}

The proposed Light-ML structure achieves feature lifting within the student model. Similar to the channel shuffle in ShuffleNet~\citep{ShufflenetV12018}, Light-ML augments features by integrating information between different task branches during forward propagation, while in backward propagation, it optimizes the model from the detection heads to the neck using gradient and distillation information from both branches as additional supervisory clues, which further improve the overall utilization of supervision information, thereby alleviating inadequate training due to insufficient distillation information and enhancing the distillation process. Therefore, the Light-ML structure could alleviate the insufficient supervision issue caused by low foreground ratio, thus suitable in drone imagery detection.

\subsection{Centerness-based Instance-aware Distillation}
\label{sec:method_cid}

As illustrated in Sec.\ref{sec:method_preliminaries}, information weighting results $\mathbb{I}$ highlight more consequential knowledge to distill and improving the accuracy of the training procedure. LD~\citep{ld2022} analyzes knowledge distribution patterns related to classification and localization tasks, identifying context regions of instances as Valuable Localization Regions (VLR) for localization distillation. 

However, despite the fact that the localization branch heavily relies on information around small instances in drone image detection, previous work~\citep{ld2022} suffers from the low confidence issue due to the low DIoU value when capturing valuable localization information, leading to unreliable knowledge, thus inappropriate for drone images. 

Specifically, previous methods propose VLR based on regions where the DIoU value is lower than $\gamma_{LD}\cdot \alpha_{pos}$, setting the distillation weight to IoU. Here, $\gamma_{LD}$ serves as a hyper-parameter, and $\alpha_{pos}$ as the threshold for positive sample region. In drone imagery detection, small instances lead to low IoU value between the ground truth (GT) boxes and the predefined anchor boxes. As a result, previous method could not provide adequate supervision due to the low distillation weight, thus inappropriate for drone imagery.

Moreover, lower DIoU values lead the previous methods to set a lower positive sample threshold $\alpha_{pos}$, resulting in the expansion of positive sample regions, which causes VLR to deviate from small instances. Consequently, the teacher model distills more background noise instead of valuable localization knowledge.

To address them, we propose a novel centerness-based instance-aware distillation technique, aimed at introducing a size-robust metric to enhance distillation efficiency.

To effectively identify potential valuable regions around the instances, we introduce the centerness target~\citep{FCOS2019} for weighting purposes. For the $i$-th level, predefined anchors $\bm{A}_i\in \mathbb{R}^{n\times 2}$ and corresponding bounding boxes $\bm{\mathcal{B}}_i \in \mathbb{R}^{n\times 4}$ is accomplished through detection heads. The centerness can be derived as follows:
\begin{equation}
{centerness}_i=\sqrt{\frac{\min(l_i^*, r_i^*)}{\max(l_i^*, r_i^*)}\times\frac{\min(t_i^*, b_i^*)}{\max(t_i^*, b_i^*)}}
\end{equation}
where $l_i^*$, $r_i^*$, $t_i^*$ and $b_i^*$ come from location offsets calculated through $\bm{A}_i$ and $\bm{\mathcal{B}}_i$.

Considering the properties of centerness, it is solely relying on the position relative to the corresponding bounding boxes, making it robust to size variations. Additionally, when using the GT boxes as $\bm{\mathcal{B}}_i$, as in FCOS~\citep{FCOS2019}, the centerness will exhibit high values in the positive sample regions of small instances and low values in the contextual regions.

Nevertheless, in FCOS, centerness is only defined within the GT boxes, which may lead to insufficient supervision, since the low foreground ratio in drone imagery. Therefore, we utilize the detection results derived from fully trained teacher models as the basis for information weighting.

Specifically, we employ the predicted bounding boxes or the integral results of the predicted regression distributions from trained teacher models as $\bm{\mathcal{B}}_i\in \mathbb{R}^{n\times 4}$ to derive centerness. Additionally, we calculate the diagonal lengths of each GT boxes, and apply 0.75 times these lengths as the thresholds to preliminary filter the distillation regions. This approach allows our centerness target to be size-robust, effectively indicating whether the position lies within the contextual regions surrounding the small instances.

Since our primary objective is to capture valuable information surrounding instances, we prioritize the selection of anchors $\bm{A}_i$ characterized by lower centerness values, which signify greater deviation between predicted objects and these anchors, indicating their proximity to the contexts. Therefore, the centerness-based instance-aware region weighting can be derived as follows:
\begin{equation}
    \mathbb{I}_{vlr}=
    \begin{cases}
       
        1 - centerness_i, & centerness_i< \gamma\\
        0,                & centerness_i \geq \gamma\\
       
    \end{cases}
    \label{eq:w_vlr}
\end{equation}
where the hyper-parameter $\gamma \in [0,1]$ controls region weighting degree, the coverage of valuable regions will gradually expand with $\gamma$ increases.

Combining centerness-based instance-aware regions $\mathbb{I}_{vlr}$ with positive regions $\mathbb{I}_{main}$ of the positive samples generated from the label assignment process for positive samples, the comprehensive focal distillation within the localization branch can be formulated as:
\begin{equation}
    \mathcal{L}^{dis}_{f}=\frac{1}{N}\sum_i{(\mathbb{I}_{main}+\alpha \mathbb{I}_{vlr}){L}_{reg}(\bm{\mathcal{B}}^T_i, \bm{\mathcal{B}}^S_i)}
\label{eq:l_focal}
\end{equation}
where $\alpha$ denoted the loss weighting of centerness-based instance-aware regions. 

Compared to the VLR proposed in LD~\citep{ld2022}, the centerness-based instance-aware region only depends on its position in the predicted box, robust to small size, making it more general and avoiding the low DIoU problem. Furthermore, it places more emphasis on the regions around the small instances, which could effectively focus on local information.

As merely distilling knowledge within the positive sample regions and their immediate contexts can be insufficient in transferring enough supervisory information from the teacher model, especially concerning drone images with a significantly lower foreground ratio. Therefore, we additionally adopt a global distilling approach for features proposed by~\citep{mgd2022} within the localization branch, which incorporates a Masked Image Modeling technique for global information reconstruction on the student model. The masked feature $\bm{M}_i$ is set to $1$ at each pixel with a probability determined by the hyper-parameter $\lambda$. And the reconstruction module $\mathcal{G}$ is lightweight and consists of two convolutional layers. The overall global distillation loss can be obtained from:
\begin{equation}
    \mathcal{L}^{dis}_g=\frac{1}{N} \sum_i {{L}({\bm{\mathcal{B}}}^T_i, \mathcal{G}(\bm{\mathcal{B}}^S_i\times \bm{M}_i))}
\end{equation}
where ${L}$ refers to the Smooth-L1 loss.

\begin{table*}[ht]
    \centering
   
    \caption{Comparsion of AP/AR (\%) and GFLOPs on VisDrone by our proposed method and existing state-of-the-art methods, with GFL-ResNet101 as teacher and GFL-ResNet50/GFL-ResNet18 as students.}
   
    \resizebox{0.8\linewidth}{!}{
        \begin{minipage}{\linewidth}
        \centering
        \fontsize{13.5pt}{15pt}\selectfont
        \setlength{\parindent}{0pt}
        \setlength{\tabcolsep}{5pt}{
            \begin{tabular}{l|ccc|cccc|c}
            \toprule
           
            Method & mAP & $\text{AP}_{50}$ & $\text{AP}_{75}$ & $\text{AR}_{1}$ & $\text{AR}_{10}$ & $\text{AR}_{100}$ & $\text{AR}_{500}$ & GFLOPs \\ \midrule
            GFL-ResNet101 (T) & 29.7 & 52.0 & 29.4 & 0.59 & 6.51 & 37.0 & 44.0 & 294.2 \\
            GFL-ResNet50  (S) & 24.6 & 45.2 & 23.4 & 1.06 & 7.88 & 32.0 & 37.8 & 214.3 \\
           
            FGD~\citep{FGD2022}           & 26.7 & 48.1 & 25.8 & \textbf{1.12} & \textbf{8.10} & 35.1 & 39.0 & 214.3 \\

            LD~\citep{ld2022}            & 28.9 & 50.6 & 28.6 & 0.64 & 6.52 & 36.4 & 43.2 & 214.3 \\
           
            \rowcolor{lightgray!45}LD + \textbf{Ours}     & 29.2 & 51.1 & 29.0 & 0.72 & 6.58 & 36.8 & 43.8 & 217.6 \\
           
            BCKD~\citep{bckd2023}          & 29.5 & \textbf{52.4} & \textbf{29.2} & 0.59 & 6.25 & 36.7 & 43.7 & 214.3 \\
           
            \rowcolor{lightgray!45}BCKD + \textbf{Ours}   & \textbf{29.7} & 52.1 & \textbf{29.2} & 0.61 & 6.39 & \textbf{37.0} & \textbf{43.9} & 217.6 \\ \midrule
           
            GFL-ResNet101 (T) & 29.7 & 52.0 & 29.4 & 0.59 & 6.51 & 37.0 & 44.0 & 294.2 \\
            GFL-ResNet18  (S) & 23.4 & 43.6 & 22.0 & \textbf{1.06} & \textbf{7.52} & 31.2 & 37.5 & 162.3 \\

            FGD~\citep{FGD2022}           & 25.7 & 46.3 & 24.9 & 0.82 & 7.28 & 33.8 & 38.1 & 162.3 \\
           
            LD~\citep{ld2022}            & 26.5 & 47.0 & 25.8 & 0.61 & 6.19 & 34.1 & 41.0 & 162.3 \\
           
            \rowcolor{lightgray!45}LD + \textbf{Ours}     & 27.8 & 49.6 & 27.3 & 0.78 & 6.48 & 35.2 & 42.4 & 165.6 \\
           
            BCKD~\citep{bckd2023}          & 27.9 & \textbf{49.8} & 27.4 & 0.60 & 6.21 & 35.4 & 42.5 & 162.3 \\
           
            \rowcolor{lightgray!45}BCKD + \textbf{Ours}   & \textbf{28.2} & \textbf{49.8} & \textbf{27.9} & 0.61 & 6.23 & \textbf{35.9} & \textbf{42.8} & 165.6 \\ \bottomrule

            \end{tabular}
        }
        \end{minipage}
    }
   
    \label{tab:sota1}
\end{table*}

\begin{table*}[ht]
    \centering
    \caption{Comparsion of mAP/AP (\%) and GFLOPs on UAVDT by our proposed method and existing state-of-the-art methods, with GFL-ResNet101 as teacher and GFL-ResNet50/GFL-ResNet18 as students.}
   
    \resizebox{0.8\linewidth}{!}{
        \begin{minipage}{\linewidth}
        \centering
        \fontsize{13.5pt}{15pt}\selectfont
        \setlength{\parindent}{0pt}
        \setlength{\tabcolsep}{5pt}{
            \begin{tabular}{l|ccc|c}
            \toprule
            Method & mAP & $\text{AP}_{50}$ & $\text{AP}_{75}$ & GFLOPs \\ \midrule
            GFL-ResNet101 (T) & 20.0 & 34.5 & 21.2 & 152.6 \\
            GFL-ResNet50  (S) & 19.0 & 31.3 & 22.0 & 111.4 \\
           
            FGD~\citep{FGD2022}           & 19.0 & 32.5 & 20.6 & 111.4 \\
           
            LD~\citep{ld2022}            & 20.4 & 33.2 & 23.2  & 111.4 \\
           
            \rowcolor{lightgray!45}LD + \textbf{Ours}     & \textbf{21.3} & \textbf{35.2} & \textbf{24.2} & 113.1 \\
            BCKD~\citep{bckd2023}          & 20.0 & 33.8 & 22.1 & 111.4 \\
           
            \rowcolor{lightgray!45}BCKD + \textbf{Ours}   & 20.9 & 35.0 & 23.4 & 113.1 \\ \midrule
           
            GFL-ResNet101 (T) & 20.0 & 34.5 & 21.2 & 152.6 \\
            GFL-ResNet18  (S) & 16.6 & 30.0 & 17.3 & 84.4 \\
           
            FGD~\citep{FGD2022}           & 15.6 & 28.4 & 15.6 & 84.4 \\
           
            LD~\citep{ld2022}            & 18.0 & 30.4 & 19.9 & 84.4 \\
           
            \rowcolor{lightgray!45}LD + \textbf{Ours}     & 18.5 & 31.2 & \textbf{20.5}& 86.1 \\
           
            BCKD~\citep{bckd2023}          & 18.2 & \textbf{31.9} & 19.1 & 84.4 \\
           
            \rowcolor{lightgray!45}BCKD + \textbf{Ours}   & \textbf{18.6} & 31.7 & 20.4 & 86.1 \\ 
           
            \bottomrule                           
            
            \end{tabular}
        }
        \end{minipage}
    }
   
    \label{tab:sota4}
\end{table*}

Combining the focal loss $\mathcal{L}^{dis}_f$ with the global loss $\mathcal{L}^{dis}_g$, the overall distillation loss on the localization branch can be formulated as:
\begin{equation}
    \mathcal{L}^{dis}_{loc}=\mathcal{L}^{dis}_f + \beta \mathcal{L}^{dis}_g
\label{eq:l_loc}
\end{equation}
where $\beta$ donated the weighting of global distillation loss.

\begin{table*}[ht]
    \centering
    \caption{Comparision of mAP/AP (\%) and GFLOPs on COCO between our proposed method and existing state-of-the-art methods, with GFL-ResNet101 as teacher and GFL-ResNet50/GFL-ResNet18 as students.}
   
    \resizebox{0.8\linewidth}{!}{
        \begin{minipage}{\linewidth}
        \centering
        \fontsize{13.5pt}{15pt}\selectfont
        \setlength{\parindent}{0pt}
        \setlength{\tabcolsep}{5pt}{
            \begin{tabular}{l|ccc|ccc|c}
            \toprule
            Method & mAP & $\text{AP}_{50}$ & $\text{AP}_{75}$ & $\text{AP}_{\text{S}}$ & $\text{AP}_{\text{M}}$ & $\text{AP}_{\text{L}}$ & GFLOPs \\ \midrule
            GFL-ResNet101 (T) & 44.9 & 63.1 & 49.0 & 28.0 & 49.1 & 57.2 & 294.2 \\
            GFL-ResNet50  (S) & 40.1 & 58.2 & 43.1 & 23.3  & 44.4 & 52.5 & 214.3 \\
           
            FGD~\citep{FGD2022}           & 41.3 & 58.8 & 44.8 & 24.5 & 45.6 & 53.0 & 214.3 \\
           
            LD~\citep{ld2022}            & 42.1 & 60.3 & 45.6 & 24.5 & 46.2 & 54.8 & 214.3 \\
           
            \rowcolor{lightgray!45}LD + \textbf{Ours}     & 43.0 & 60.7 & 46.9 & 25.5 & 46.9 & 55.8 & 217.6 \\
           
            BCKD~\citep{bckd2023}          & 43.2 & 61.6 & 46.9 & 25.7 & \textbf{47.3} & 55.9 & 214.3 \\
           
            \rowcolor{lightgray!45}BCKD + \textbf{Ours}   & \textbf{43.3} & \textbf{61.9} & \textbf{47.3} & \textbf{26.1} & \textbf{47.3} & \textbf{56.1} & 217.6 \\ \midrule
            GFL-ResNet101 (T) & 44.9 & 63.1 & 49.0 & 28.0 & 49.1 & 57.2 & 294.2 \\
            GFL-ResNet18  (S) & 35.8 & 53.1 & 38.2 & 18.9 & 38.9 & 47.9 & 162.3 \\
           
            FGD~\citep{FGD2022}           & 38.3    & 55.4    & 41.4    & 21.3    & 41.5 & 50.8 & 162.3 \\
           
            LD~\citep{ld2022}            & 37.5 & 54.7 & 40.4 & 20.2 & 41.2 & 49.4 & 162.3 \\
           
            \rowcolor{lightgray!45}LD + \textbf{Ours}     & 38.6 & 55.4 & 41.9 & 21.5 & 41.5 & 50.8 & 165.6 \\
           
            BCKD~\citep{bckd2023}          & 38.6 & 56.4 & 41.7 & 21.4 & 42.0 & 50.0 & 162.3 \\
           
            \rowcolor{lightgray!45}BCKD + \textbf{Ours}   & \textbf{39.6} & \textbf{57.5} & \textbf{42.7} & \textbf{22.2} & \textbf{42.6} & \textbf{51.6} & 165.6 \\ \bottomrule                           
            
            \end{tabular}
        }
        \end{minipage}
    }
   
    \label{tab:sota3}
\end{table*}

\begin{table*}[ht]
    \centering
    \caption{Comparsion of AP/AR (\%) and GFLOPs on VisDrone by our proposed method and existing state-of-the-art methods on different dense object detectors (FCOS and ATSS).}
   
    \resizebox{0.8\linewidth}{!}{
        \begin{minipage}{\linewidth}
        \centering
        \fontsize{13.5pt}{15pt}\selectfont
        \setlength{\parindent}{0pt}
        \setlength{\tabcolsep}{5pt}{
            \begin{tabular}{l|ccc|cccc|c}
            \toprule
                Method & mAP & $\text{AP}_{50}$ & $\text{AP}_{75}$ & $\text{AR}_{1}$ & $\text{AR}_{10}$ & $\text{AR}_{100}$ & $\text{AR}_{500}$ & GFLOPs \\
            \midrule
                FCOS-ResNet101 (T) & 27.4    & 48.7    & 26.6    & 0.53    & 6.61    & 35.3    & 43.3 & 294.2 \\
                FCOS-ResNet18  (S) & 22.6    & 42.1    & 21.3    & \textbf{0.65}    & \textbf{6.48}    & 30.8    & 37.9 & 162.3 \\

                FGD~\citep{FGD2022} & 23.6 & 42.8 & 22.5 & 0.51 & 5.80 & 31.6 & 36.7 & 162.3 \\
               
                LD~\citep{ld2022} & 25.4    & 45.6    & 24.6    & 0.51    & 6.07    & 33.3    & 41.3 & 162.3 \\
               
                \rowcolor{lightgray!45}LD + \textbf{Ours}     & 25.9    & 46.4    & 24.9    & 0.58    & 6.23    & 33.9    & 41.9 & 165.6 \\
               
                BCKD~\citep{bckd2023} & 26.5    & 47.8    & 25.5    & 0.53    & 6.29    & 34.5    & 42.6 & 162.3 \\
               
                \rowcolor{lightgray!45}BCKD + \textbf{Ours}   & \textbf{26.8}    & \textbf{48.1}    & \textbf{26.0}    & 0.57   & 6.35    & \textbf{34.8}    & \textbf{42.7} & 165.6 \\ 
               
            \midrule
                    ATSS-ResNet101 (T) & 28.2 & 49.2 & 28.0 & 0.66 & 6.66 & 36.1 & 43.8 & 294.2 \\
                    ATSS-ResNet18  (S) & 23.7 & 43.4 & 22.8 & \textbf{0.72} & 6.77 & 31.6 & 38.5 & 162.3 \\

                    FGD~\citep{FGD2022}           & 25.6 & 45.4 & 25.2 & 0.64 & \textbf{7.00} & 34.1 & 38.5 & 162.3 \\

                    LD~\citep{ld2022}            & 26.0 & 45.8 & 25.8 & 0.60 & 6.30 & 34.1 & 41.8 & 162.3 \\
                   
                    \rowcolor{lightgray!45}LD + \textbf{Ours}     & 26.6 & 46.8 & 26.3 & 0.70 & 6.58 & 34.6 & 42.9 & 165.6 \\
                   
                    BCKD~\citep{bckd2023}          & 27.2 & \textbf{48.0} & 26.9 & 0.63 & 6.49 & 35.1 & 42.9 & 162.3 \\

                    \rowcolor{lightgray!45}BCKD + \textbf{Ours}   & \textbf{27.4} & \textbf{48.0} & \textbf{27.2} & 0.63 & 6.35 & \textbf{35.3} & \textbf{43.1} & 165.6 \\
                   
            \bottomrule                                                    
            \end{tabular}
        }
        \end{minipage}
    }
   
    \label{tab:sota2}
\end{table*}

As the Centerness-based Instances-aware Distillation (CID) focuses on the distillation region weighting, our proposed method exhibits adaptability for integration into various prevailing state-of-the-art methodologies by replacing the loss function $L_{reg}$ in a flexible way.

\section{Experimental Results and Analysis}

In this section, we evaluate the effectiveness of our proposed knowledge distillation approach by comparing it to the state-of-the-art distillation methods on various datasets and conducting extensive ablation studies.

\subsection{Datasets and Evaluation Metrics}

We conduct experiments on three benchmarks widely employed in drone object detection and general object detection tasks, \ie VisDrone~\citep{VisDrone2018}, UAVDT~\citep{UAVDT2018}, and COCO~\citep{COCO2014}. \textbf{VisDrone} contains 7,019 high-resolution drone images with 10 categories. Following previous work~\citep{ClusDet2019,ufpmpdet2022,ceasc2023}, we select 6,471 images for training and 548 images for testing. \textbf{UAVDT} contains 23,258 training images and 15,069 test images with 3 categories and a resolution of 1024$\times$540 pixels. \textbf{COCO} contains 118,000 images for training and 5,000 for testing with 80 categories.

\begin{table*}[ht]
    \centering
    \caption{Ablation study on components with GFL ResNet101-ResNet18.}
   
    \resizebox{0.75\linewidth}{!}{
        \begin{minipage}{\linewidth}
        \centering
        \fontsize{13.5pt}{15pt}\selectfont
        \setlength{\parindent}{0pt}
        \setlength{\tabcolsep}{5pt}{
            \begin{tabular}{c|cc|ccc|c}
            \toprule
            Dataset & Light-ML & CID and Global & mAP & $\text{AP}_{50}$ & $\text{AP}_{75}$ & GFLOPs \\ 
                \midrule
                   \multirow{4}{*}{VisDrone} & & & 26.5 & 47.0 & 25.8 &162.3 \\
                  
                   & \checkmark & & 27.0 & 47.9 & 26.3 & 165.6 \\
                   & & \checkmark & 27.2 & 48.6 & 26.6 & 162.3 \\
                  
                   & \checkmark & \checkmark & \textbf{27.8} & \textbf{49.6} & \textbf{27.3} & 165.6 \\ 
                \midrule
                    \multirow{4}{*}{COCO} &&& 37.5 & 54.7 & 40.4 &162.3\\
                     &\checkmark & & 38.4 & 55.1 & 41.8 &165.6\\
                     && \checkmark & 38.0 & \textbf{55.4} & 41.0 & 162.3 \\
                     &\checkmark & \checkmark & \textbf{38.6} & \textbf{55.4} & \textbf{41.9} & 165.6\\
            \bottomrule
                                                    
            \end{tabular}
        }
        \end{minipage}
    }
   
    \label{tab:ab1}
\end{table*}

We report Average Precision (AP) and Average Recall (AR) as the evaluation metrics for accuracy, and GFLOPs as the metric for efficiency.

\subsection{Implementation Details}

We implement the proposed method using PyTorch~\citep{pytorch2019} and MMDetection~\citep{mmdetection2019}. All the student models on the VisDrone and COCO datasets are trained for 12 epochs with SGD optimizer. The learning rate is initially set at 0.01 with linear warm-up strategy and decreases by 10 after 8 and 11 epochs. On UAVDT dataset, we train models for 6 epochs with an initial learning rate of 0.01, and decreases by 10 after 4 and 5 epochs. The input image sizes are set to 1333$\times$800 for VisDrone and COCO, and 1024$\times$540 for UAVDT. The hyperparameters $k$ in Sec.~\ref{sec:method_light-ml} and $\gamma$ in Eq.~\eqref{eq:w_vlr} are set as $\{k=0.25, \gamma=0.45\}$, and following the previous work, the hyperparameters $\alpha$ in Eq.~\eqref{eq:l_focal}, $\lambda$ in Sec.~\ref{sec:method_cid}, and $\beta$ in Eq.~\eqref{eq:l_loc} are set as $\{\alpha=1, \lambda=0.65, \beta=4\}$. The training phase for VisDrone and UAVDT is conducted on 2 Nvidia RTX 2080Ti GPUs, and 8 Nvidia RTX 2080Ti GPUs on COCO dataset.

\subsection{Main results}

Our methods can be applied to various logit-based distillation approaches, thus we apply our method to LD~\citep{ld2022} and BCKD~\citep{bckd2023} methods respectively. The performance of these models is evaluated by comparing them with the state-of-the-art distillation methods, including FGD~\citep{FGD2022}, LD, and BCKD. 

As reported in Table~\ref{tab:sota1}, we evaluate performance using GFL~\citep{GFLv12020} as the base detector and ResNet101 as the teacher model, while employing ResNet50 and ResNet18 as the student models on VisDrone dataset. Our method significantly surpasses the feature-based method FGD, exceeding it by over 3.0\% and 2.5\% on different students, and achieves considerable improvements of 0.2\%-1.3\% in different basic logit-based methods, with only a marginal increase in computational cost by 3.3 GFLOPs. It is important to note that the ResNet101-ResNet50 distillation, based on BCKD and our method, promotes the student model's mAP accuracy to 29.7\%, equaling that of the teacher model.

\begin{table*}[ht]
    \centering
    \caption{Ablation study on detailed designs on Light-ML with GFL ResNet101-ResNet18 on VisDrone.}
    \resizebox{0.7\linewidth}{!}{
        \begin{minipage}{\linewidth}
        \centering
        \fontsize{13.5pt}{15pt}\selectfont
        \setlength{\parindent}{0pt}
        \setlength{\tabcolsep}{5pt}{
            \begin{tabular}{l|cc|ccc|c}
            \toprule
            Method            & Cls to Reg & Reg to Cls & mAP & $\text{AP}_{50}$ & $\text{AP}_{75}$ & GFLOPs \\ \midrule
            Baseline          &&& 27.2 & 48.6 & 26.6 & 162.3 \\
           
            Channel Shuffle~\citep{ShufflenetV12018}   &\checkmark&\checkmark& 27.6 & 49.2 & 27.1 & 162.3 \\
           
            Cls to Reg Fusion~\citep{GFLv22021}          &\checkmark&& 27.2 & 48.4 & 26.6 & 163.8 \\
            Reg to Cls Fusion~\citep{GFLv22021}          &&\checkmark& 27.0 & 48.1 & 26.5 & 162.7 \\
            \rowcolor{lightgray!45} \textbf{Light-ML (Ours)}              &\checkmark&\checkmark& \textbf{27.8} & \textbf{49.6} & \textbf{27.3} & 165.6 \\ \bottomrule                
                                                    
            \end{tabular}
        }
        \end{minipage}
    }
   
    \label{tab:ab2}
\end{table*}

\begin{table*}[ht]
    \centering
    \caption{Ablation on the hyper-parameter $k$ with GFL ResNet101-ResNet18 on VisDrone.}
    \resizebox{0.7\linewidth}{!}{
        \begin{minipage}{\linewidth}
        \centering
        \fontsize{13.5pt}{15pt}\selectfont
        \setlength{\parindent}{0pt}
        \setlength{\tabcolsep}{5pt} {
            \begin{tabular}{c|ccc|c}
            \toprule
            hyper-parameter $k$ & mAP & $\text{AP}_{50}$ & $\text{AP}_{75}$ & GFLOPs \\ \midrule
            0 & 27.6 & 49.2 & 27.1 & \textbf{162.3} \\
            0.25 & \textbf{27.8} & \textbf{49.6} & \textbf{27.3} & 165.6 \\
            0.50 & \textbf{27.8} & 49.4 & 27.2 & 175.5 \\
            0.75 & \textbf{27.8} & 49.3 & 27.1 & 191.9 \\
            1    & 27.7 & 49.0 & 27.0 & 214.9 \\ \bottomrule   
            \end{tabular}
        }
        \end{minipage}
    }
    \label{tab:suppab2}
\end{table*}

On another benchmark for drone object detection UAVDT, as reported in Table~\ref{tab:sota4}, our method significantly surpasses FGD by over 2.3\% and 3.0\% in mAP on various student models, and shows an improvement of 0.4\%-0.9\% compared to basic methods, indicating the generalizability of our approach.

We also conduct corresponding experiments on the popular COCO dataset for general object detection. As summarized in Table~\ref{tab:sota3}, our method achieves improvements of 2.0\% and 1.3\% in various student models compared to the feature-based method FGD. Additionally, it shows improvements ranging from 0.1\%-1.1\% in basic logit-distillation methods, with only a minor increase in computational demand. 

It is noteworthy that, as shown in Table~\ref{tab:sota3}, when employing ResNet18 as the student model, we observe a significant relative improvement of 6.4\% and 3.7\% compared to the baseline in the accuracy of $\text{AP}_{\text{S}}$, greater than those in $\text{AP}_{\text{M}}$ (0.7\% and 1.4\%) and $\text{AP}_{\text{L}}$(2.8\% and 3.2\%). It demonstrates the effectiveness of our newly proposed Light-ML model on weaker student models and the impact of CID on smaller instances, indicating that our contributions are beneficial for the student to learn from the teacher in locating small instances which are dominant on drone images.

To further demonstrate the generalizability of our method, we further validate our method on other base detectors, \ie FCOS~\citep{FCOS2019} and ATSS~\citep{ATSS2020}, as shown in Table~\ref{tab:sota2}, employing ResNet101 as the teacher model and ResNet18 as the student model on VisDrone dataset. Similar to GFL, our method shows an improvement of 1.8\%-3.2\% in mAP over the feature-based method FGD, and 0.2\%-0.6\% over basic logit-based methods, demonstrating the flexibility of our distillation approach.

\subsection{Ablation Studies}

\subsubsection{On Light-ML and CID}

As shown in Table~\ref{tab:ab1}, the adoption of the Light-ML component leads to a 0.5\%-1.0\% improvement in mAP performance on VisDrone and COCO datasets with only an additional 2\% in GFLOPs, largely due to the reduced performance gap between teacher and student models, as well as enhanced utilization of supervisory information from the Light-ML structure. Adopting the localization branch distillation method CID and Global distillation leads to a 0.7\% and 0.5\% improvement, primarily attributed to the superior adaptability of our proposed method to small instances. By combining Light-ML with CID and Global distillation, the performance improvement further increases to 1.3\% and 1.0\% respectively  on VisDrone and COCO, which demonstrates the remarkable efficacy of the proposed method. 

Significantly, the adoption of the Light-ML component and our distillation method results in a 4.9\% relative improvement on the VisDrone dataset, surpassing the 2.9\% relative improvement achieved on COCO. This further demonstrates the efficacy of our method in drone imagery detection.

\begin{table*}[ht]
    \centering
    \caption{Ablation study of CID and Global distillation with GFL ResNet101-ResNet18 on VisDrone.}
    \resizebox{0.7\linewidth}{!}{
        \begin{minipage}{\linewidth}
        \centering
        \fontsize{13.5pt}{15pt}\selectfont
        \setlength{\parindent}{0pt}
        \setlength{\tabcolsep}{5pt}{
            \begin{tabular}{cc|ccc|c}
            \toprule
            CID & Global Distillation & mAP & $\text{AP}_{50}$ & $\text{AP}_{75}$ & GFLOPs \\ \midrule
           
             & & 27.0 & 47.9 & 26.3 & 165.6 \\
             \checkmark & & 27.7 & 49.1 & \textbf{27.3} & 165.6 \\
              & \checkmark & 27.0 & 47.9 & 26.2 & 165.6 \\
             \checkmark & \checkmark & \textbf{27.8} & \textbf{49.6} & \textbf{27.3} & 165.6 \\ \bottomrule
                                                    
            \end{tabular}
        }
        \end{minipage}
    }
    \label{tab:rebuttal-3}
   
\end{table*}

\begin{figure*}[ht]
    \centering
    \resizebox{0.99\linewidth}{!}{
        \subfloat[]{
            \centering
            \includegraphics[width=0.33\linewidth,keepaspectratio]{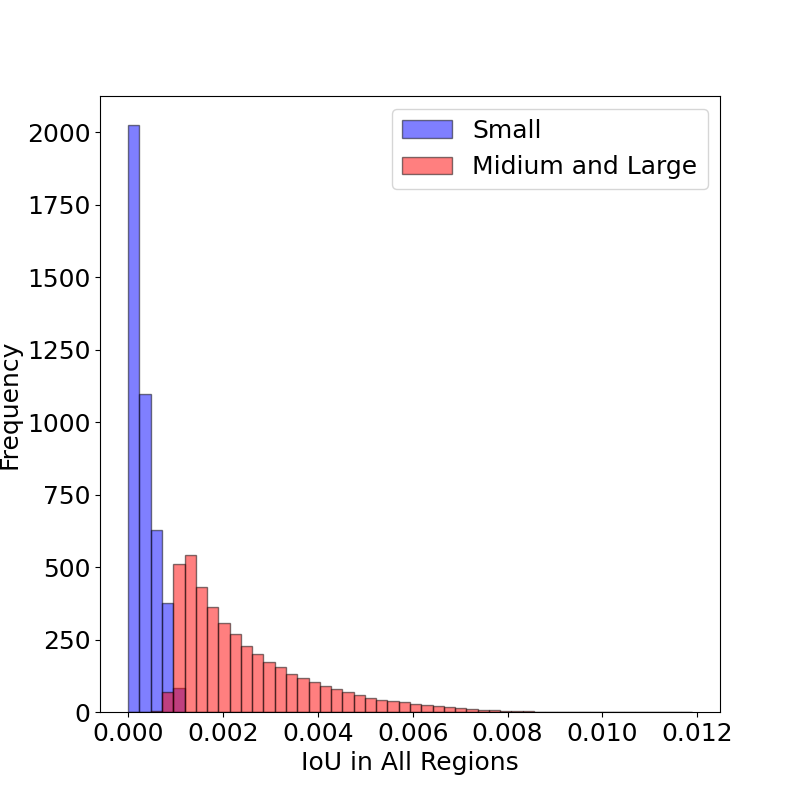}
        }
        \subfloat[]{
            \centering
            \includegraphics[width=0.33\linewidth,keepaspectratio]{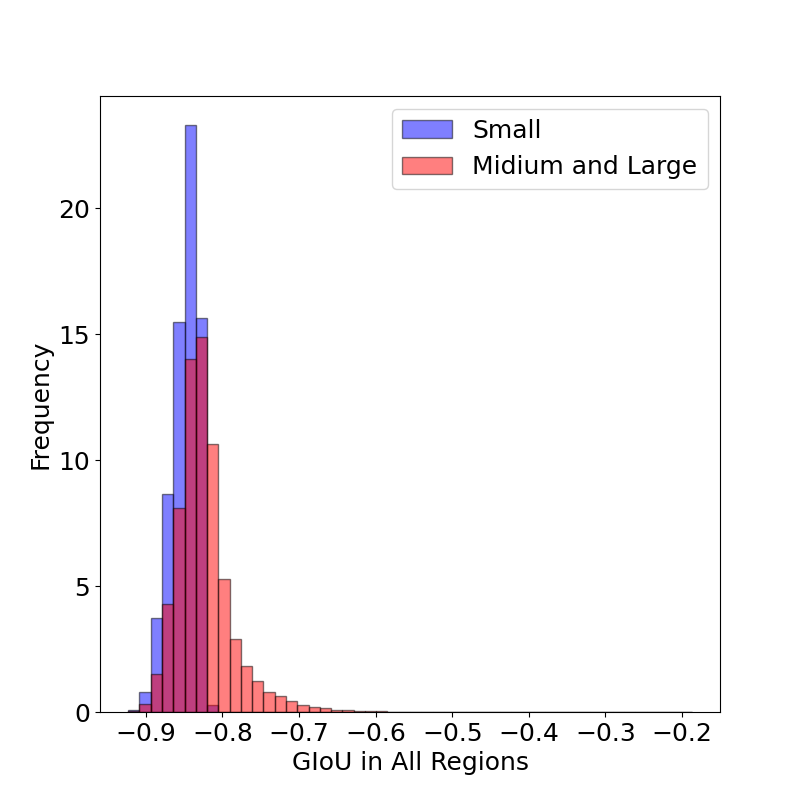}
        }
        \subfloat[]{
            \centering
            \includegraphics[width=0.33\linewidth,keepaspectratio]{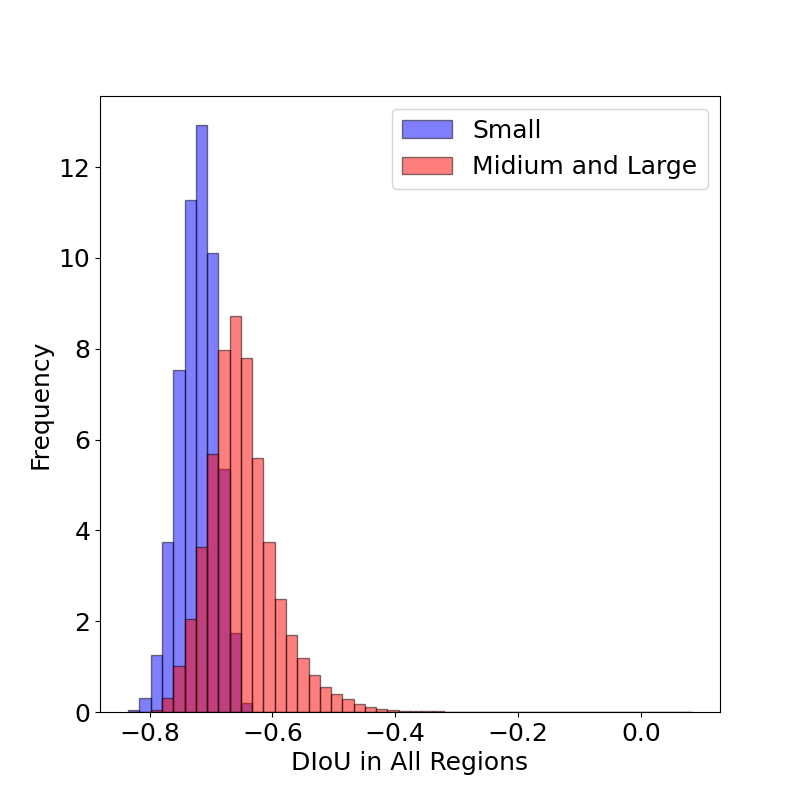}
        }
    }
    \caption{Visualization of the frequency distribution of the mean (a) IoU, (b) GIoU and (c) DIoU value for each ground truth box in the VisDrone dataset, where small instances refer to instances with an area less than $32\times 32$} 
   
    \label{fig:iou_distribution}
   
\end{figure*}

\begin{table*}[ht]
    \centering
    \caption{Ablation on the hyper-parameter $\gamma$ with GFL ResNet101-ResNet18 on VisDrone.}
   
    \resizebox{0.7\linewidth}{!}{
        \begin{minipage}{\linewidth}
        \centering
        \fontsize{13.5pt}{15pt}\selectfont
        \setlength{\parindent}{0pt}
        \setlength{\tabcolsep}{5pt}{
            \begin{tabular}{c|cccccc}
            \toprule
            $\gamma$ & 0.6 & 0.55 & 0.5 & 0.45 & 0.4 & 0.35 \\ 
           
            \midrule
            mAP(\%) & 27.4 & 27.6 & 27.7 & \textbf{27.8} & 27.7 & 27.5\\
           
            \bottomrule
            \end{tabular}
        }
        \end{minipage}
    }
    \label{tab:result}
   
\end{table*}

\subsubsection{On Light-ML}
\label{sec:ablation_on_lightml}

We evaluate the performance of our proposed Light-ML structure by comparing it with other lightweight feature lifting structures, including channel shuffle~\citep{ShufflenetV12018} for all channels, feature fusion from the classification branch to the regression branch, as well as from the regression branch to the classification branch, using an attention-based approach proposed in ~\citep{GFLv22021}, denoted as "Cls to Reg Fusion" and "Reg to Cls Fusion", respectively. As displayed in Table~\ref{tab:ab2}, the two attention-based fusion methods show a performance decline, primarily attributed to the incomplete feature lifting from different tasks. The application of channel shuffling results in a 0.4\% performance improvement, whereas our approach achieves a superior performance-efficiency trade-off by introducing the CSP structure and additional convolution, leading to a 0.6\% improvement in accuracy with only a 2\% additional GFLOPs. 

As described in Sec.\ref{sec:method_light-ml}, $k$ signifies the division ratio of convolution and channel shuffle operators. The computational cost of the Light-ML module grows as $k$ increases. In Table~\ref{tab:suppab2}, we report the detection accuracy and efficiency for various values of $k$. The results indicate that the student model achieves the best trade-off between accuracy and efficiency when $k$ = 0.25, which is therefore used in our experiments.

\subsubsection{On CID}

\begin{figure*}[ht]
    \centering
    \resizebox{0.66\linewidth}{!}{
   
        \subfloat[]{
            \centering
            \includegraphics[width=0.33\textwidth,keepaspectratio]{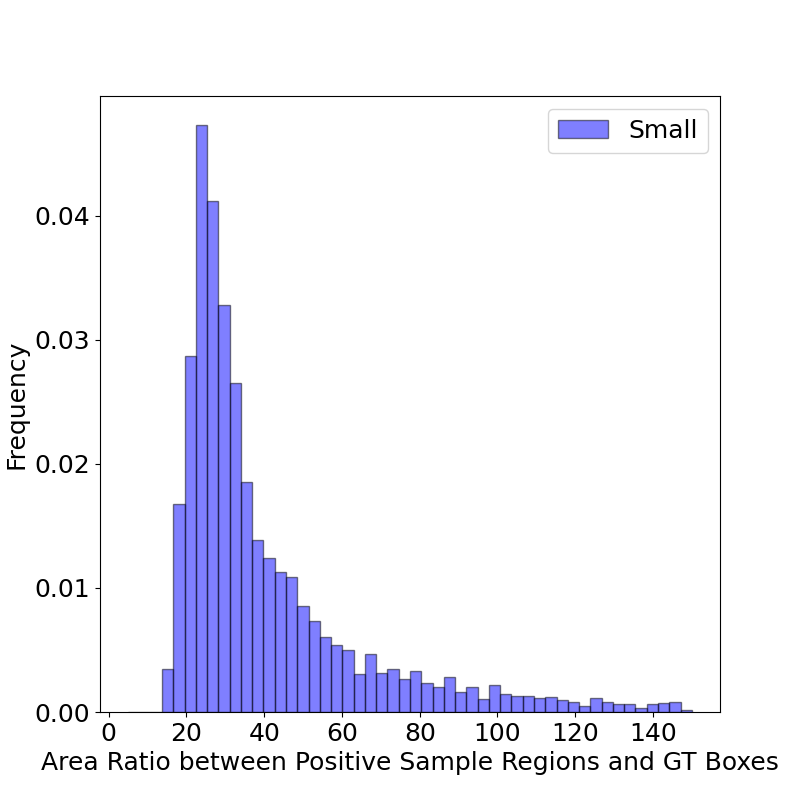}
        }
        \subfloat[]{
            \centering
            \includegraphics[width=0.33\textwidth,keepaspectratio]{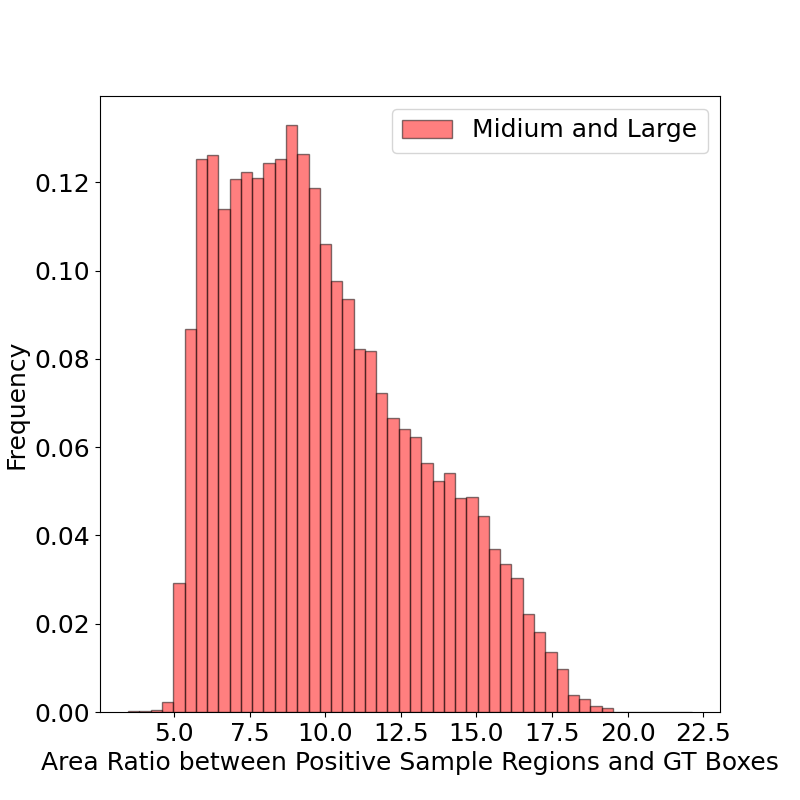}
        }

    }
    \caption{Visualization of the frequency distribution of the area ratio between the positive sample regions of LD and the ground truth boxes in the VisDrone dataset, where: (a) instances with an area less than $32\times 32$, (b) instances with an area not less than $32\times 32$.}

    \label{fig:positive_gt_frequency}
\end{figure*}

\begin{figure*}[ht]
    \centering
    \resizebox{1.0\linewidth}{!}{
        \includegraphics[width=1.0\linewidth,keepaspectratio]{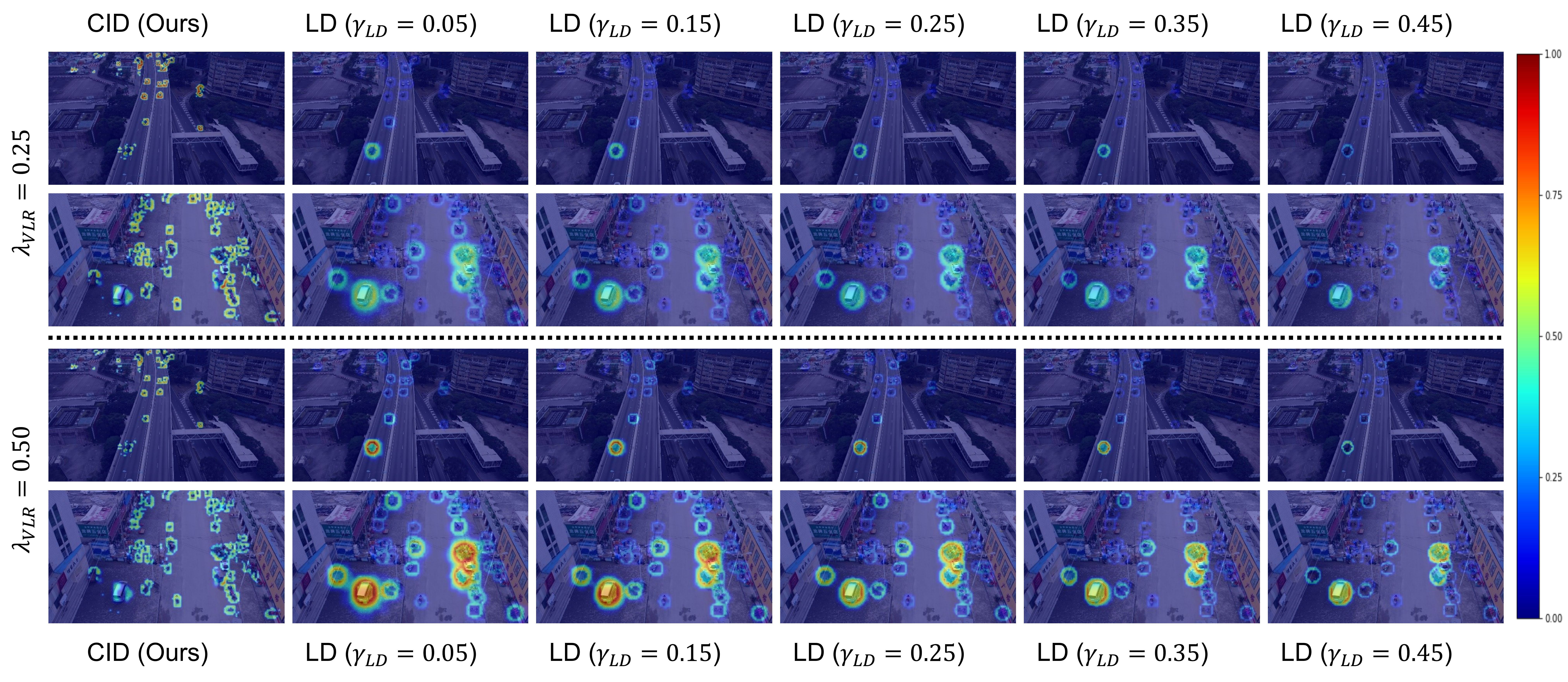}
       
    }
    \caption{Visualization of VLR regions using CID and LD with different $\gamma_{LD}$ values. Highlighted areas indicate activated regions for distillation.} 
   
    \label{fig:cidldvlr_1x}
   
\end{figure*}

\begin{figure*}[ht]
    \centering
    \resizebox{0.6\linewidth}{!}{
        \includegraphics[width=0.6\linewidth,keepaspectratio]{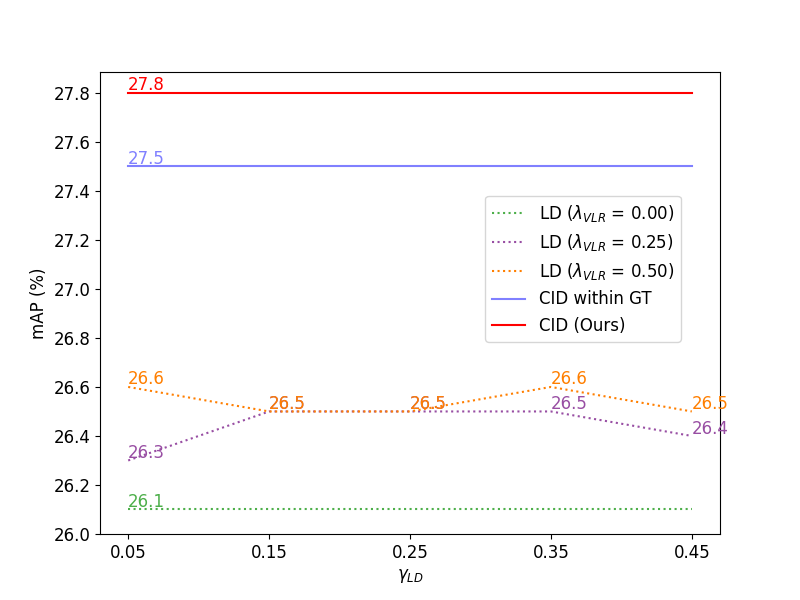}
    }
    \caption{Ablation study on the different VLR regions and weights in GFL ResNet18 on VisDrone. "LD ($\lambda_{VLR}$=0.00/0.25/0.50)" denote LD with different distillation weight on VLR; "CID within GT" indicates the use of the FCOS method for calculating centerness.}
    \label{fig:cid-ld-gamma}
\end{figure*}

\begin{figure*}[!t]
    \centering
    \resizebox{0.9\linewidth}{!}{
        \includegraphics[width=1.0\linewidth,keepaspectratio]{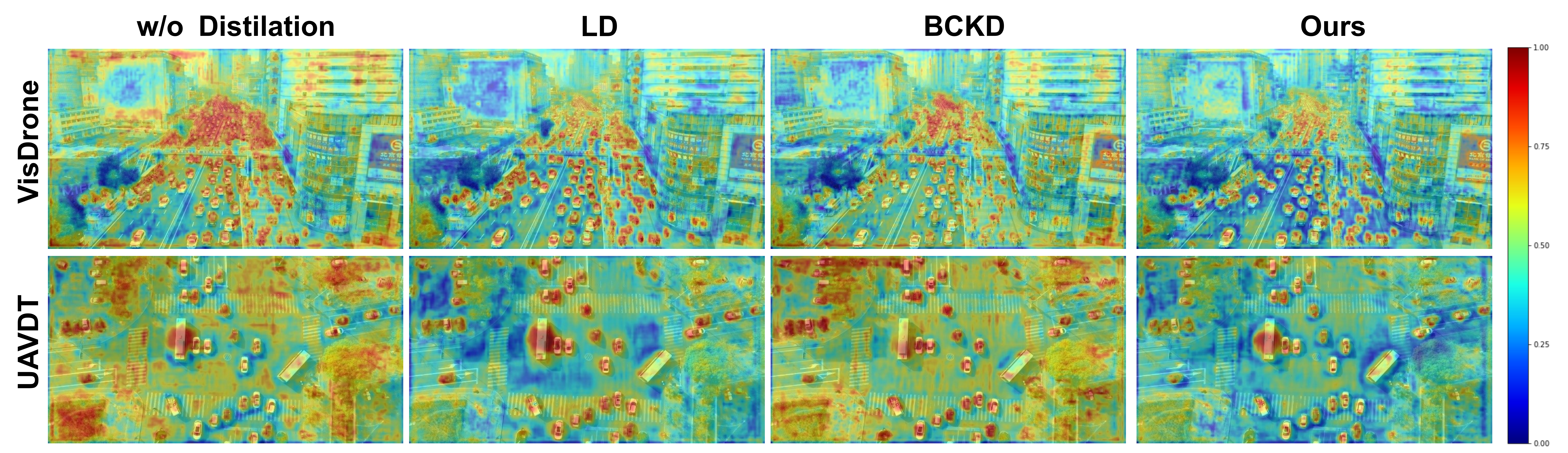}
    }
    \caption{Visualization of cosine distance of the localization branch between the teacher (GFL-ResNet101) and the student (GFL-ResNet18) at the P3 level of the FPN. Lower cosine distance indicates better performance.} 
    \label{fig:vis}
   
\end{figure*}

We separately evaluate the effect of Centerness-based Instance-aware Distillation and Global Distillation in Table~\ref{tab:rebuttal-3}. When employing the CID method, student models can achieve a better distillation effect around small instances, resulting in a 0.7\% improvement in accuracy. Moreover, only utilizing global distillation results in inferior performance as it cannot provide sufficient clues around instances. However, when we combine the global distillation with CID, it further enhances the utilization of potential distillation information in background region, leading to a 0.1\% improvement in accuracy. 

As described in Eq.\eqref{eq:l_focal}, $\gamma$ influences the extent of region weighting in the CID method. In Table~\ref{tab:result}, we report the detection accuracy for various values of $\gamma$. The results indicate that the student model achieves the highest performance when $\gamma$=0.45, which is therefore used in our experiments.

As illustrated in Fig.~\ref{fig:iou_distribution}, we compile statistics on the IoU, GIoU and DIoU distribution corresponding to each ground truth box. It is obvious that the IoU, GIoU and DIoU values for the ground truth of small instances are relative lower than those of medium and large instances. This demonstrate that the small instances are constrained by their smaller instances, leading to a lower distillation weight in LD, which in turn results in LD being unable to distill adequate supervisory information.

To demonstrate that LD introduces more background noise in small instances, we calculate the frequency distribution of the area ratio between the positive sample regions of LD and the ground truth boxes. As shown in Fig.~\ref{fig:positive_gt_frequency}, LD sets a larger positive sample region for smaller instances, resulting in a farther VLR, which in turn introduces more background noise.

We also visualize the VLR proposed by LD with different $\gamma_{LD}$ values and the VLR proposed by our CID. As shown in Fig.~\ref{fig:cidldvlr_1x}, LD generates a farther VLR for small instances with a lower distillation information weight $\mathbb{\bm{I}}_{feat}$. Moreover, although a lower $\gamma_{LD}$ assigns a higher $\mathbb{\bm{I}}_{feat}$ to regions far from the small instances, compared to our CID, this $\mathbb{\bm{I}}_{feat}$ is still insufficient to distill enough knowledge. In contrast, CID generates a more precise VLR with a higher distillation information weight, demonstrating the effectiveness of our method.

To address the issue of low distillation information weight, a common practice is to assign a larger weight to the VLR in the loss function. Therefore, we double the distillation weight in the VLR proposed by LD with different $\gamma_{LD}$ values. As shown in Fig.~\ref{fig:cidldvlr_1x}, the doubled distillation information weights $\mathbb{\bm{I}}_{feat}$ show some advantage over the weight used in CID. 

Therefore, we double the distillation weight hyper-parameter $\lambda_{VLR}$ of LD on VLR from the default 0.25 to 0.5, and conduct experiments on LD with different $\gamma_{LD}$. However, as shown in Fig.~\ref{fig:cid-ld-gamma}, LD still struggles to distill more knowledge to the student model, but it remains inferior to CID due to the background noise.

Furthermore, when we block the information in the VLR during distillation, the student model still achieves an mAP accuracy of 26.1, which is close to the mAP of LD. This demonstrates that the VLR proposed by LD is far from the small instances, introducing background noise rather than valuable localization knowledge. In contrast, our CID method proposes the VLR around small instances, making it more suitable for drone imagery detection.

It is worth noting that when we adopt the centerness calculation similar to that used in FCOS, to propose the VLR and set the distillation weight to $1-{centerness}_i$, the student model achieves an improvement of 0.9\%. This result further confirms that LD suffers the issue that it distills background noise, making it unsuitable in drone imagery detection. However, as illustrated in Fig.~\ref{fig:cid-ld-gamma}, the traditional centerness method still falls short in distillation due to the low ratio of foreground regions in drone imagery. Consequently, we extend the concept of centerness to be computed across the entire feature maps, leading to an additional improvement of 0.3\%. This demonstrates the effectiveness of our method.

\subsubsection{Visualization of Knowledge Distillation}

To demonstrate the effectiveness of our distillation method on the localization branch, we visualize the performance distance between the teacher and student models at the P3 level of the FPN for small instances on the VisDrone and UAVDT datasets. As shown in Fig.\ref{fig:vis}, our approach significantly reduces the performance gap between the teacher and student models in the positive sample regions and the VLR regions surrounding the instances important for localization, thereby confirming the effectiveness of our method.

\section{Conclusion}

We propose a novel knowledge distillation method for drone image object detection. It first introduces a Light-ML structure designed to improve the utilization of supervision information during distillation by task-wise feature lifting, boosting the distillation efficiency consequently. Meanwhile, it proposes a localization branch distillation strategy integrating Centerness-based Instance-aware Distillation to refine the distillation around the small instances. Extensive experiment results conducted across VisDrone, COCO, and UAVDT datasets demonstrate the improvement in accuracy with competitive computation on our proposed approach.

\section*{Data Availability Statements}
The VisDrone~\citep{VisDrone2018}, UAVDT~\citep{UAVDT2018} and COCO~\citep{COCO2014} databases used in this manuscript are deposited in publicly available repositories respectively: \url{https://github.com/VisDrone/VisDrone-Dataset}, \url{https://sites.google.com/view/grli-uavdt} and \url{https://cocodataset.org}.

\bibliographystyle{spbasic}     
\bibliography{egbib}  

\end{document}